\documentclass[12pt,a4paper]{cibb}

\makeatletter
\providecommand{\@ordinalM}[2]{#1}
\makeatother

\usepackage{subfigure,graphicx}
\usepackage{amsmath,amsfonts,latexsym,amssymb,euscript,xr}
\usepackage{booktabs}
\usepackage[nodayofweek]{datetime}
\usepackage{hyperref}
\usepackage{fmtcount}
\usepackage[english]{datenumber}
\usepackage[absolute]{textpos}

\usepackage{algorithm}
\usepackage{algorithmic}
\usepackage{float}

\usepackage[table]{xcolor}
\usepackage{color,colortbl,tabularx}

\usepackage[english]{babel}
\usepackage[protrusion=true,expansion=true]{microtype}
\usepackage{amsmath,amsfonts,amsthm}
\usepackage{pifont}

\def\blue{\color{blue}}

\definecolor{LightBlue}{rgb}{0.88,0.9,0.9}

\title{\Large $\ $\\ \bf Quantum Kernel Estimation for the
Discovery of Early Lung Cancer Detection}

\author{\blue \large Hamed Javidi$^1$, Alex Zajichek$^{2}$, Hakan Doga$^{3}$, Laxmi Parida$^{3}$, Filippo Utro$^{*,3}$ and Peter J. Mazzone$^{*,4}$}

\address{\blue \footnotesize $\ $\\$^1$ Department of Computational Life Sciences, Cleveland Clinic Research, Cleveland, Ohio, USA. \\
$^2$ Department of Quantitative Health Sciences, Cleveland Clinic Research,
Cleveland, Ohio, USA. \\
$^3$ IBM Quantum, IBM Thomas J Watson Research Center, Yorktown Heights, NY, USA.  \\
$^4$ Pulmonary Department, Cleveland Clinic, Cleveland, Ohio, USA. \\

$^*$corresponding author: futro@us.ibm.com and mazzonp@ccf.org
}

\abstract{\small quantum machine learning, quantum kernel estimation, cfDNA fragmentomics, DNA methylation, lung cancer. \normalsize
\\[17pt]

{\bf Abstract.} Lung cancer screening with low-dose chest computed tomography reduces mortality, but its
impact is limited by uptake, adherence, and management challenges. Blood-based cell-free DNA
(cfDNA) biomarkers offer a complementary approach, although early detection remains difficult
because of lung cancer heterogeneity and high-dimensional, nonlinear molecular signals. We
evaluated quantum-classical hybrid machine learning for lung cancer detection using DNA
fragmentomics and DNA methylation. After feature selection, models were trained using 20- and
40-feature subsets. Features were encoded into quantum Hilbert space using angle and dense-angle
feature maps with multiple entanglement strategies. Fidelity-based quantum kernels
were computed with exact statevector simulation and integrated with precomputed-kernel SVM
and kernel-PCA logistic regression and compared with an SVM model trained on the original features.

This framework enabled systematic evaluation of how encoding and entanglement design affect
classification. Across repeated held-out evaluations, quantum-kernel models achieved competitive
performance on both datasets. For fragmentomics, several 20-feature configurations improved
AUC relative to a classical SVM baseline, suggesting effective capture of nonlinear cfDNA
fragmentation structure. For methylation, the classical SVM achieved the highest AUC, although
selected quantum models remained competitive and improved specificity in some cases.
Increasing features from 20 to 40 did not consistently improve performance and often increased
variability. Overall, these results support quantum kernel methods as a promising approach for
cfDNA-based lung cancer detection.
}
\begin{document}

\renewcommand{\thefootnote}{}
\footnotetext{\small{Article version: \datedate $\;$ h\currenttime  $\;$ CET}}

\thispagestyle{myheadings}
\pagestyle{myheadings}
\markright{\tt Proceedings of CIBB 2026}

\section{Introduction}
\label{sec:SCIENTIFIC-BACKGROUND}

Lung cancer screening with an annual low-dose chest CT scan (LDCT) has been shown to reduce mortality among individuals at elevated risk who are suitable candidates for early-stage intervention\cite{nlst2011reduced,dekoning2020nelson}. However, the overall impact of screening programs has been attenuated by slow uptake among eligible populations, low adherence to annual screening, and channelnges in management of  screen detected findings, such as indeterminate pulmonary nodules. 

More accurate, accessible, and lower-risk diagnostic tools could help to improve the implementation of lung cancer screening and help optimize the balance between clinical benefits and harm at the population level. In this context, blood based molecular biomarkers, including cell-free DNA (cDNA) assay are being explored. However, lung cancer is a biologically heterogeneous disease arising in diverse patient populations, which complicates the identification of robust and generalizable molecular signature. Existing  biomarkers approaches oftern exhibit a trad-off between sensitivity and specificity, limiting their clinical utility. 

Classical machine learning models may struggle by the high dimensionality and complex potentially nonlinear interactions present in molecular data.  Quantum and quantum-classical hybrid machine learning approaches offer an alternative framework for encoding and processing such data, with the potential to capture richer feature representations.  In this study, we evaluate quantum-classical hybrid models for early lung cancer detection using two cfDNA biomarker datasets: a fragmentomics-based assay derived from~\cite{mathios2021delfi,mazzone2024fragmentome}, and a methylation-based signature from a  the second a discovery-level study study.

\section{Data and Methods}
\label{sec:DATA-AND-METHODS}

Two datasets derived from studies aimed at developing blood based tests for lung cancer screening were analized.

The first dataset consists of a cfDNA methylation-based assay. This dataset includes methylation measurements of 56 genomic targets, identified by a methylation-sensitive restriction endonuclease approach. The cohort comprises  813 individuals enrolled in a case-control biomarker development study (currently unpublished). All participants met eligible criteria for standard of care lung cancer screening based on age and smoking history. Among them, 188 (23\%) individuals had lung cancer, of whom 150 (80\%) were stage I-III. The mean age was 67 years.

The second is a publicly available cDNA  fragmentomics dataset~\cite{mathios2021delfi}. It includes fragment lengths profiles from 718 individuals. Of these, 365 individual with concerning symptoms or suspicious findings on chest imaging were used as the training cohort in the original study. The remaining samples were obtained from participants enrolled in colorectal cancer screening trials, supplemented with early-stage lung cancer cases from a biorepository. In particular, the dataset includes 172 (24\%) individuals with lung cancer, of whom 99 (58\%) were stage I-III. The mean age of the cohort was 60 years. Low-coverage whole-genome sequencing was used to derive cfDNA fragment features, computed as ratios of short to long fragments across 473 non-overlapping 5 MB regions (each containing approximately 80,000 fragments), spanning 2.4 GB of the genome.


In both dataset, all cancers (stages I-IV) and controls were included in the final analyses. Feature selection was performed using L1-penalized logistic regression\cite{hastie2009elements}. Within the training data, features were standardized and a logistic regression model with L1 regularization was fitted using stratified cross-validation, with the regularization parameter selected to maximize the area under the receiver operating characteristic curve (AUC). 
Features were ranked by the absolute value of their model coefficients, and the top-ranked 20 or 40 features were retained for downstream  modeling. These selected feature were subsequently used as inputs for both classical and quantum-kernel methods. 

\subsection{Quantum Kernel Estimation}

We implemented a quantum kernel estimation (QKE) framework to construct nonlinear similarity matrices from classically selected features. In QKE, classical inputs are embedded into a quantum Hilbert space via a parameterized feature map, and sample similarity is defined by the fidelity between the resulting quantum states \cite{schuld2019quantum,havlicek2019supervised}. This formulation induces an implicit nonlinear kernel though the quatum circuit, while downstream classification is performed using classical methods.

Let $\mathbf{x}_i \in \mathbb{R}^{d}$ denote the selected feature vector for sample $i$, with $d \in \{20,40\}$. Features were scaled to $[0,\pi/2]$ using statistics computed on the training-set:
\begin{equation}
    \tilde{x}_{ij}
    =
    \frac{x_{ij}-\min(x_j)}
    {\max(x_j)-\min(x_j)}
    \cdot \frac{\pi}{2}.
\end{equation}
The scaling parameters were fitted exclusively on the training set and subsequently applied to validation and held-out test data to prevent information leakage.

Two quantum encoding strategies were evaluated independently. In angle encoding, one feature was mapped to each qubit, yeielding $n_q=d$. In dense-angle encoding, two features were assigned per qubit, yielding $n_q=d/2$. Each scaled feature vector was then encoded into a quantum state $
    |\phi(\tilde{\mathbf{x}})\rangle
    =
    U_{\phi}(\tilde{\mathbf{x}};\theta)
    |0\rangle^{\otimes n_q}$, and $
    \theta=\frac{\pi}{2}.
$
The feature map consisted of a global $R_y(\theta)$ layer, followed by a controlled-$Z$ entangling layer, and data-dependent sinlge-qubit rotations. Three entanglement maps were considered as separate experiments configurations: (i) correlation-sorted linear, (ii) custom scoring-based, and (iii) shuffled linear connectivity. Correlation-sorted linear entanglement uses Pearson correlation to determine feature order within a fixed linear qubit chain, whereas custom scoring-based entanglement uses Pearson-correlation-derived scores to directly select the strongest qubit-pair connections. As a control, shuffled linear entanglement uses the same fixed linear chain after randomly shuffling the feature order, allowing us to assess whether correlation-informed ordering improves the quantum kernel representation.

For angle encoding, the feature map was
\begin{equation}
    U_{\mathrm{angle}}(\tilde{\mathbf{x}};\theta)
    =
    \left[
    \prod_{q=1}^{n_q}
    R_z^{(q)}(-2\tilde{x}_{q})
    \right]
    \left[
    \prod_{(q,r)\in \mathcal{E}} CZ_{q,r}
    \right]
    \left[
    \prod_{q=1}^{n_q} R_y^{(q)}(\theta)
    \right].
\end{equation}
For dense-angle encoding, the feature map was
\begin{equation}
    U_{\mathrm{dense}}(\tilde{\mathbf{x}};\theta)
    =
    \left[
    \prod_{q=1}^{n_q}
    R_z^{(q)}(-2\tilde{x}_{2q+1})
    R_x^{(q)}(-2\tilde{x}_{2q})
    \right]
    \left[
    \prod_{(q,r)\in \mathcal{E}} CZ_{q,r}
    \right]
    \left[
    \prod_{q=1}^{n_q} R_y^{(q)}(\theta)
    \right].
\end{equation}

The quantum kernel between samples was defined as the squared state overlap of their corresposning quantum states $
    K(\mathbf{x}_i,\mathbf{x}_j)
    =
    \left|
    \langle
    \phi(\tilde{\mathbf{x}}_i)
    |
    \phi(\tilde{\mathbf{x}}_j)
    \rangle
    \right|^2$.
This fidelity kernel is the standard quantum kernel used in quantum feature-space learning and fidelity-based quantum kernel classifiers \cite{schuld2019quantum,havlicek2019supervised}. 
In this work, kernels were computed using exact statevector simulation, which enables precise estimation of the quantum feature-space similarity. Extension to quantum hardware would involve finite sampling and device noise, which may affect kernel quality and will be investigated in future work~\cite{hubregtsen2022training}.

Let $\mathbf{S}_{\mathrm{train}}$ and $\mathbf{S}_{\mathrm{test}}$ denote matrices of simulated statevectors for training and test sample, respectively. Kernel matrices were computed as $
    \mathbf{K}_{\mathrm{train}}
    =
    \left|
    \mathbf{S}_{\mathrm{train}}
    \mathbf{S}_{\mathrm{train}}^{\dagger}
    \right|^{\circ 2},
    \qquad
    \mathbf{K}_{\mathrm{test}}
    =
    \left|
    \mathbf{S}_{\mathrm{test}}
    \mathbf{S}_{\mathrm{train}}^{\dagger}
    \right|^{\circ 2}.
$
The training kernel was symmetrized, its diagonal was set to one, and all entries were clipped to $[0,1]$ to mitigate numerical instabilities.

The precomputed quantum kernels were passed to two downstream classifiers: an SVM 
trained directly on $\mathbf{K}_{\mathrm{train}}$, and logistic regression on a kernel 
principal component analysis (KPCA) embedding, with data split 80/20 into training and 
held-out test sets. As a classical comparator, a separate SVM was trained on raw 
features to mirror the SVM used with the quantum kernel, isolating kernel construction 
(classical vs.\ quantum) as the sole variable of comparison; its kernel type (linear, 
RBF, polynomial, sigmoid) and hyperparameters were jointly optimized via Optuna (TPE 
sampler) to maximize mean AUC under stratified 5-fold cross-validation.

\begin{figure}[t]
\centering
\includegraphics[width=1\textwidth]{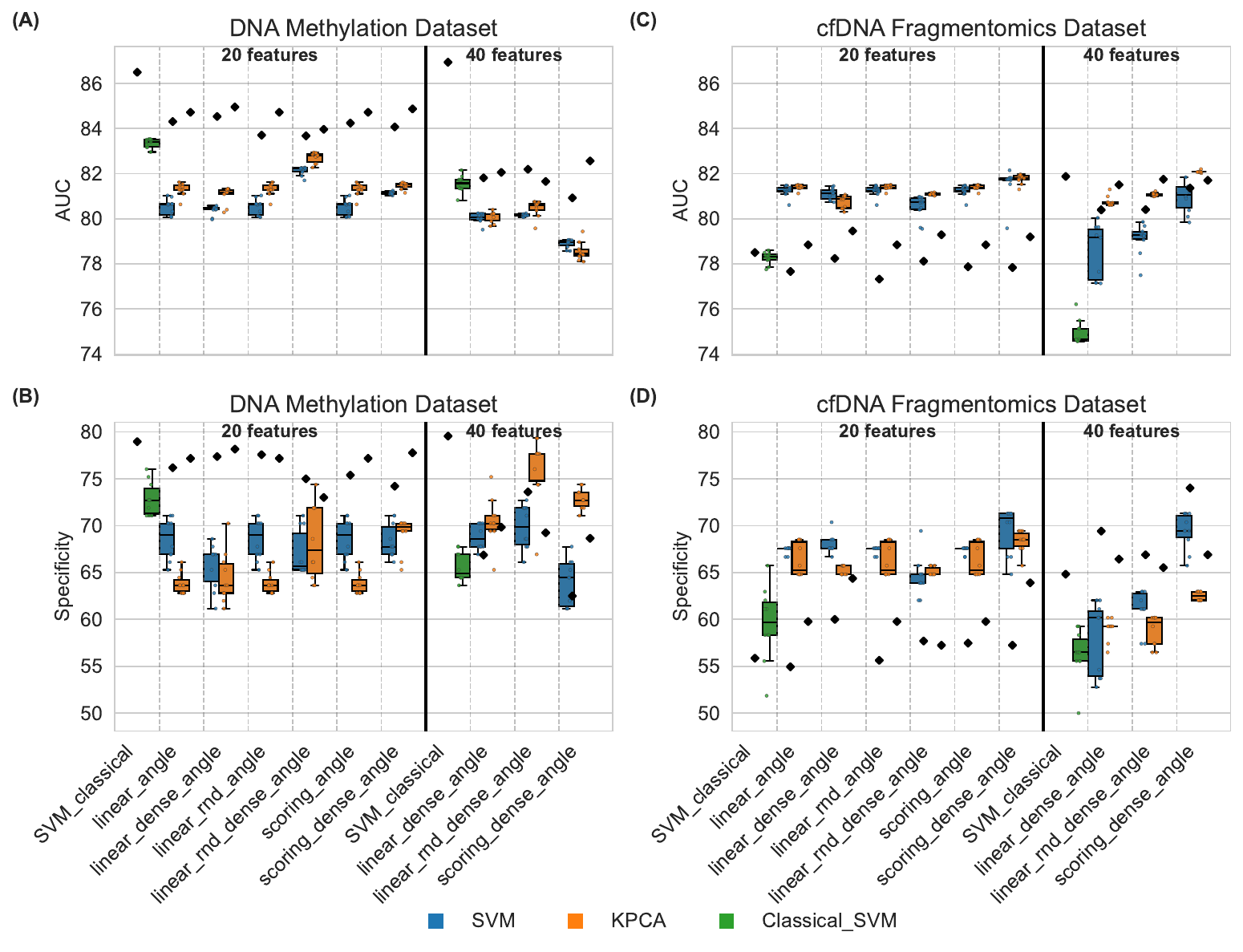}
\caption{
\textbf{Held-out and cross-validation performance}
AUC and specificity at a fixed sensitivity of 80\% are shown for classical SVM, quantum-kernel SVM, and QPCA models across feature-selection, encoding, and entanglement configurations. Boxplots summarize repeated held-out test performance over 10 runs, while black diamond markers indicate cross-validation estimates.
}
\label{fig:results}
\end{figure}

\section{Results}
\label{sec:RESULTS}

\subsection{DNA methylation Dataset}

The performance comparison on the DNA methylation dataset is shown in Fig.~\ref{fig:results}A-B. Overall, the classical SVM baseline using the top 20 features achieved the highest held-out AUC among the evaluated models, with values centered around approximately 83--84\%. Quantum-kernel models, including both the SVM and KPCA classifiers, achieved competitive but generally lower AUC values, typically ranging from approximately 80--82.5\% across the 20-feature configurations.

Within the 20-feature quantum experiments, KPCA-based models generally achieved slightly higher AUC than the precomputed-kernel SVM, particularly for both angle and dense-angle encoding schemes. Among these, the dense-angle linear configuration yielded one of the strongest AUC performances results. In contrast, increasing the selected features  from 20 to 40 did not consistently improve performance. Several 40-feature dense-angle configurations showed reduced AUC, in some cases decreasing to approximately 79--80\%, suggesting that the additional features may induced redundancy, noise, or less stable quantum similarity structure.

Specificity exhibited greater variability than AUC across repeated held-out evaluations, as shown in Fig.~\ref{fig:results}B. The classical SVM with 20 selected features achieved specificity values centered around approximately 72--74\%, whereas the 40-feature variant showed reduced specificityapproximately 65--67\%. Among quantum-kernel models, KPCA showed relatively strong specificity in selected 40-feature dense-angle configurations, reaching approximately 75--78\%, although with increased variability. Precomputed-kernel SVM models generally showed moderate specificity, often ranging 65--71\%, depending on the encoding and entanglement configuration.

Overall, these results indicate that quantum-kernel models can achieve performance competitive with classical baselines, although their effectiveness is sensitive to feature selection, encoding strategy, and entanglement design. The lack of consistent improvement when increasing feature dimensionality suggests that larger feature sets do not necessarily translate into better generalization for quantum kernel methods in this setting.

\subsection{cfDNA fragmentomics Dataset}

The corresponding results for the DNA fragmentomics dataset are shown in Fig.~\ref{fig:results}C-D. In contrast to the methylation dataset, quantum-kernel models achieved consistently strong AUC across several 20-feature configurations. Most 20-feature quantum experiments produced AUC values around 81--82\%, exceeding the classical SVM baseline, which was centered near approximately 78--79\%. This trend suggests that the quantum-kernel representation captured useful nonlinear structure in the fragmentomic feature space.

Across AUC results, the KPCA-based model were frequently among the best-performing approaches, particularly in the 20-feature angle and dense-angle settings. These configurations also showed relatively compact held-out distributions, indicating stable performance across repeated runs. In the 40-feature setting, performance remained competitive for some configurations, especially the scoring-based dense-angle configuration, but other showed increased variability. This effect was especially evident for the quantum-kernel SVM, where some 40-feature configurations exhibit broader AUC dispersion.

Specificity followed a similar pattern but showed greater variability than AUC across repeated held-out evalation. The 20-feature quantum-kernel models generally improved specificity relative to the classical SVM baseline, with several models reaching approximately 65--70\%. The strongest specificity values were observed for selected SVM-based quantum-kernel, particularly in dense-angle and scoring-based experiments. In contrast, the  40-feature classical SVM baseline showed reduced specificity compared to the 20-feature setting, consistent with the trend observed in the methylation dataset.

Overall, these results indicate that quantum-kernel models can improve AUC relative to the classical baseline in fragmentomics-based classification under selected configurations, particularly in the lower-dimensional (20-feature) setting. However, as observed in the methylation dataset, increasing the feature dimensionality to 40 did not consistently improve performance and often introduced additional variability.

\section{Discussion}

Across both datasets, AUC was more stable than specificity, while specificity exhibited greater variability across repeated held-out evaluations. This pattern indicates that the models were generally more consistent in ranking samples than in achieving a fixed-threshold operating point. The methylation dataset favored stronger classical SVM performance, particularly in the 20 features, whereas the fragmentomics models showed clearer improvement from quantum-kernel approaches.

The comparison between 20-feature and 40-feature configurations suggests that increasing the number of selected features did not systematically improve performance. In several cases, 40-feature models exhibited either reduced AUC or increased variability, suggesting that the lower-ranked features may introduce noise or alter the kernel geometry in a way that reduces generalization. This effect was most apparent in the methylation dataset and in selected fragmentomic SVM configurations.

Overall, three main observations emerge. First, quantum-kernel models achieved performance competitive with classical baselines across both datasets. Second, KPCA-based approaches were often more stable than precomputed-kernel SVMs in terms of AUC. Third, the 20-feature setting consistently provided a more favorable balance between predictive performance and stability than the 40-feature setting.
It is important to note, however, that the features used in these experiments were originally selected from millions of candidates using classical machine learning and statistical approaches for cancer classification. As a result, these features may be inherently biased toward classical ML methods, potentially limiting the achievable performance of quantum models. This effect is particularly evident in the methylation dataset, where only 56 sites were provided from millions of DNA methylation loci. In future work, we plan to explore feature selection strategies that are better aligned with quantum models, for example by leveraging representations that enhance non-linear structure or are more suitable for quantum downstream tasks.

\section{Conclusion}
\label{sec:CONCLUSIONS}
Overall, this work presents a proof-of-concept evaluation of quantum kernel estimation for
liquid-biopsy-based lung cancer biomarker discovery. Future work will examine alternative
feature maps, kernel designs, larger and multimodal feature sets, and adaptive kernel-learning
strategies to improve classification while preserving stability. We will also assess performance
under realistic hardware constraints, including finite sampling and noise.





\footnotesize
\bibliographystyle{unsrt}
\bibliography{bibliography_CIBB_file.bib} 
\normalsize

\end{document}